\RequirePackage[table,dvipsnames]{xcolor}
\documentclass[10pt,twocolumn,letterpaper]{article}

\usepackage[pagenumbers]{cvpr}      

\definecolor{cvprblue}{rgb}{0.21,0.49,0.74}
\usepackage[pagebackref,breaklinks,colorlinks,allcolors=cvprblue]{hyperref}

\usepackage{multirow}

\def\paperID{13132} 
\def\confName{CVPR}
\def\confYear{2025}

\title{MAKIMA: Tuning-free Multi-Attribute Open-domain Video Editing via Mask-Guided Attention Modulation}

\author{Haoyu Zheng\\
{\tt\small zhenghaoyu@zju.edu.cn}
\and
Wenqiao Zhang\\
{\tt\small wenqiaozhang@zju.edu.cn}
\and
Zheqi Lv\\
{\tt\small lvzheqi@zju.edu.cn}
\and
Yu Zhong\\
{\tt\small zhongyu@whut.edu.cn}
\and
Yang Dai\\
{\tt\small daiyang@zju.edu.cn}
\and
Jianxiang An\\
{\tt\small anjianxiang@zju.edu.cn}
\and
Yongliang Shen\\
{\tt\small shenyl@zju.edu.cn}
\and
Juncheng Li\\
{\tt\small lijuncheng@zju.edu.cn}
\and
Dongping Zhang\\
{\tt\small zhangdongping@cjlu.edu.cn}
\and
Siliang Tang\\
{\tt\small siliang@zju.edu.cn}
\and
Yueting Zhuang\\
{\tt\small yzhuang@zju.edu.cn}
}

\begin{document}
\twocolumn[{
\renewcommand\twocolumn[1][]{#1}%
\maketitle
\begin{center}
    \vspace{-1em}  
    \centering
    \captionsetup{type=figure}
    \includegraphics[width=\textwidth]{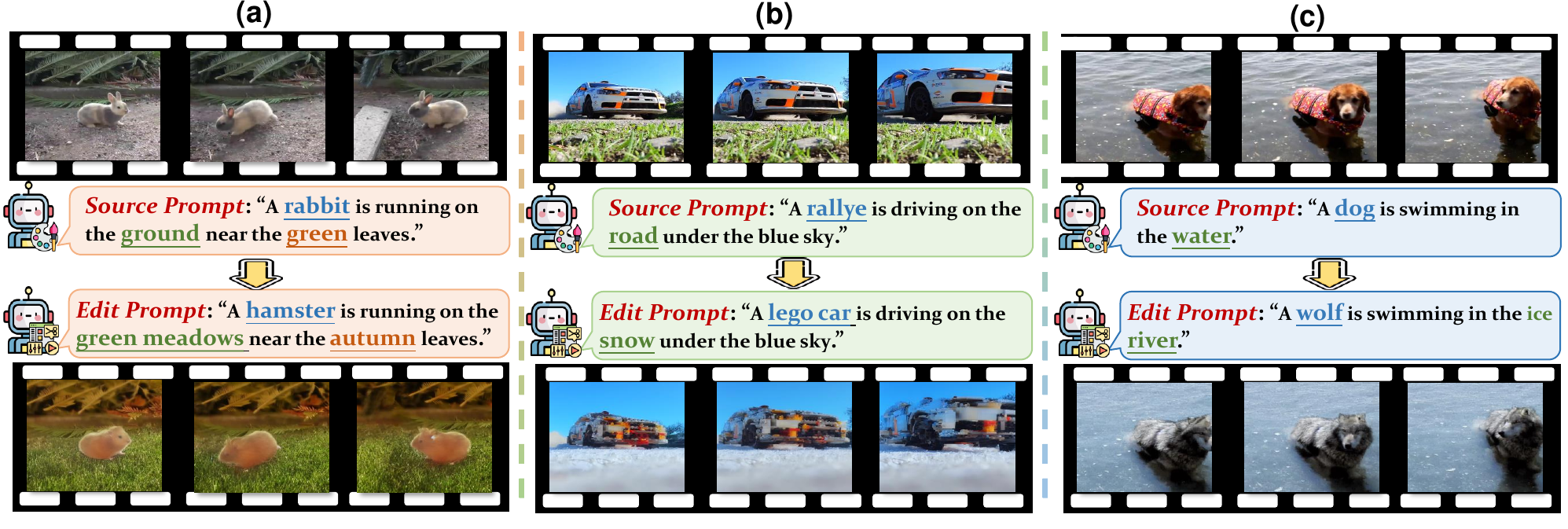}
    \captionof{figure}{MAKIMA achieves open-domain multi-attribute video editing while maintaining the structure of the source video without tuning.}
    \label{fig:teaser}

\end{center}
}]

\newcommand*{\method}{Video Editing}

\begin{abstract}

Diffusion-based text-to-image (T2I) models have demonstrated remarkable results in global video editing tasks.
However, their focus is primarily on global video modifications, achieving desired attribute-specific changes still remains a challenging task, \ie,  multi-attribute editing (MAE) in video.
Broadly, contemporary video editing approaches either necessitate extensive fine-tuning or depend on additional networks (\eg, ControlNet) for modeling multi-object appearances,  yet they remain in their infancy, offering only coarse-grained MAE solutions.
In this paper, we present MAKIMA, a tuning-free MAE framework built upon pretrained T2I models for open-domain video editing.
Our approach preserves video structure and appearance information by incorporating attention maps and features from the inversion process during denoising. 
To facilitate precise editing of multiple attributes, we introduce mask-guided attention modulation, enhancing correlations between spatially corresponding tokens and suppressing cross-attribute interference in both self-attention and cross-attention layers.
To balance video frame generation quality and efficiency, we implement consistent feature propagation, which generates frame sequences by editing keyframes and propagating their features throughout the sequence.
Extensive experiments demonstrate that MAKIMA outperforms existing baselines in open-domain multi-attribute video editing tasks, achieving superior results in both editing accuracy and temporal consistency while maintaining computational efficiency.
\end{abstract}    
\vspace{-1em}  
\section{Introduction}
\label{sec:intro}
Coupled with massive text-image datasets, text-to-image (T2I) models~\cite{ho2020denoising} demonstrating remarkable capabilities in both high-quality image synthesis, it also facilitated advances in image editing\cite{rombach2022high, ramesh2021zero, ramesh2022hierarchical}, allowing users to control various proprieties of both generated and real images. 
Along this line of research, further research~\cite{geyer2023tokenflow, qi2023fatezero} 
expand this exciting progress to the video domain,  \ie, video editing, showing the potential of T2I models for video editing applications. 
Nevertheless, current approaches mainly focus on global appearance modifications, and the progress of fine-grained local attribute editing is still lagging behind, especially the multi-attribute editing (MAE) scenarios. 

\begin{figure*}[h]
\centering
\includegraphics[width=\linewidth]{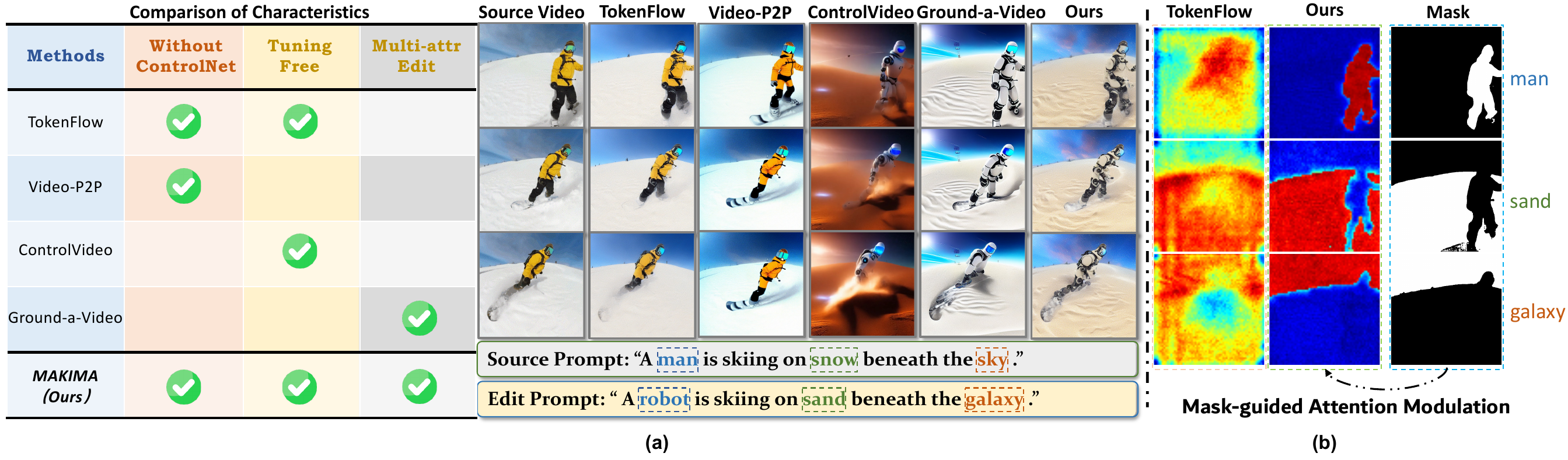}
\caption{(a) Failure cases of multi-attribute video editing by previous methods. MAKIMA achieves precise attribute modifications while preserving the structural composition of source frames. (b) Through Mask-guided Attention Modulation, MAKIMA aligns the attention distribution of different attributes with their corresponding spatial layouts in the source video.}
\label{fig:intro}
\end{figure*}

To systematically study the MAE, we begin by revisiting the contemporary video editing approaches and visualize their shortcoming for fine-grained MAE realization in Figure \ref{fig:intro} for global editing method, \eg, TokenFlow\cite{geyer2023tokenflow}, face significant limitations in achieving desired multi-attribute modifications. As a tuning-free editing method, it inherently lacks the precision needed for localized control; correspondingly, the fine-tuning approach, Video-P2P\cite{liu2024video} shares unconditional embedding but its cross-attention control becomes ineffective for some attributes during multi-attribute editing, mainly due to unreliable spatial information\cite{chefer2023attend} in the replaced attention maps; another video editing scheme, by leveraging ControlNet\cite{zhang2023adding}, \cite{zhang2023controlvideo} utilizes depth maps extracted from original frames for video editing, but this approach results in the loss of background structural information (as evident in the ``sand'' region); 
Ground-A-Video\cite{jeong2023ground}, specifically designed for multi-attribute video editing, employs word-to-bounding box control for attention modulation of different attributes. However, this mechanism encounters issues when attribute bounding boxes overlap. For instance, due to the overlap between foreground and background editing targets, the ``sand'' modification fails to materialize, and the ``robot'' exhibits structural degradation in the final frame.
Summing up, even integrating auxiliary networks or introducing cost fine-tuning, achieving finger-level MAE still remains challenging, which motivates us to reconsider the appropriate MAE solution.

In this work, to address the challenges mentioned above, we propose MAKIMA (MAsK-guided attention modulation In Multiple Attributes), which focuses on open-domain multi-attribute video editing to achieve more precise and diverse video manipulation effects. To preserve the original video frame structure without model fine-tuning while enabling detailed editing, we introduce Mutual Spatial-Temporal Self-Attention, which injects self-attention maps and features from DDIM Inversion into the denoising process, allowing the model to access structural information and appearance features from the source video frames. Previous failures in multi-attribute editing primarily stem from imprecise attention distribution, where edited objects receive insufficient attention scores in relevant regions while maintaining relatively high scores in irrelevant regions—a phenomenon known as attention leakage\cite{yang2024eva}. For self-attention, relevant regions refer to areas belonging to the same attribute across different frames, while for cross-attention, they correspond to the spatial regions associated with edited attribute text embeddings. To address this, we propose Mask-guided Attention Modulation, which enhances attention in relevant regions while suppressing it in non-relevant areas. For efficient video generation, we introduce a feature propagation mechanism that leverages both feature similarity and temporal distance. This approach enables focused editing on select keyframes with subsequent propagation to remaining frames, reducing computational overhead while maintaining visual consistency. As shown in Figure 2, extensive experiments demonstrate that our method achieves both better cost control and stronger multi-attribute video editing capabilities compared to the baselines. The improved cost control is primarily reflected in MAKIMA's avoidance of using ControlNet and fine-tuning. The enhanced multi-attribute video editing ability is demonstrated by MAKIMA's more accurate mapping of each attribute from the edit prompt onto the edited video.

Our contributions can be summarized as:
\begin{itemize}
    \item We propose MAKIMA, a novel framework for open-domain multi-attribute video editing that achieves precise manipulation using pretrained text-to-image models without any fine-tuning. 
    \item We introduce Mask-guided Attention Modulation to address the problem of imprecise attention distribution in edited objects, effectively controlling the editing process through attention enhancement and suppression.
    \item For efficient video generation, we develop a strategic feature propagation mechanism based on feature similarity and temporal distance that enables keyframe self-attention feature propagation.
    \item Without requiring model fine-tuning, our approach achieves state-of-the-art performance compared to existing baselines across various video editing tasks.
\end{itemize}

\section{Related Work}
\label{sec:related_work}
\noindent \textbf{Generation by T2I Diffusion Models.}
With the development of deep learning, image understanding\cite{zhang2022boostmis, zhang2024hyperllava} and generation technologies\cite{rombach2022high} have made remarkable advances. Early image generation relied on GANs for domain-specific distributions\cite{zhang2017stackgan, reed2016generative, li2019controllable, alaluf2021restyle}. However, these models are limited in domain generalization and editing capabilities due to their high-level feature spaces\cite{croitoru2023diffusion, yang2023diffusion}. DALL-E reformulates text-to-image (T2I) generation as sequence-to-sequence translation\cite{ramesh2021zeroshottexttoimagegeneration}, while DALL-E 2 employs CLIP\cite{radford2021learning} for text-image alignment\cite{ramesh2022hierarchical}.
Recently, diffusion-based models have shown remarkable progress\cite{tang2024any, epstein2023diffusion, huang2023diffusion, bar2024lumiere}. Denoising Diffusion Probabilistic Models (DDPMs) are widely adopted in T2I generation\cite{ho2020denoising}. Imagen\cite{ramesh2022hierarchical} introduces cascaded diffusion models with classifier-free guidance\cite{ho2022classifier}, while GLIDE\cite{nichol2021glide} enhances text conditioning through similar guidance strategies. For improved efficiency, Latent Diffusion Models (LDMs)\cite{rombach2022high} operate in compressed latent space. Built on LDMs, Stable Diffusion demonstrates exceptional generation capabilities through large-scale text-image training\cite{rombach2022high}.

\noindent \textbf{Editing by T2I Diffusion Models.}
Stable Diffusion has inspired various approaches\cite{parmar2023zero, hertz2022prompt, cong2023flatten, zhang2023controlvideo} for image editing. SDEdit\cite{meng2021sdedit} applies noise for generation, while Prompt-to-Prompt\cite{hertz2022prompt} and Pix2Pix-Zero\cite{parmar2023zero} achieve control through cross-attention. Blended Diffusion\cite{avrahami2022blended} modifies the foreground via latent blending during denoising. DDIM inversion\cite{song2020denoising} enables real image editing, while PnP\cite{tumanyan2023plug} and Masactrl\cite{cao2023masactrl} implement rigid and non-rigid editing through self-attention. \cite{kim2023dense} introduces mask-guided attention modulation, showing potential for precise video editing.
In the context of video editing, a naive frame-by-frame approach lacks temporal continuity. Tune-A-Video\cite{wu2023tune} extends to the spatiotemporal domain via one-shot tuning but struggles with local details. ControlVideo\cite{zhang2023controlvideo} utilizes depth maps and poses with ControlNet\cite{zhang2023adding}, yet lacks fine-grained control. TokenFlow\cite{geyer2023tokenflow} combines keyframe sampling with PnP\cite{tumanyan2023plug} mechanisms but suffers from limited local editing capability, resulting in ineffective feature propagation during multi-attribute editing.
FateZero\cite{qi2023fatezero} separates editing targets using cross-attention maps but suffers from attention leakage. FRESCO\cite{yang2024fresco} employs flow-guided attention yet struggles with multi-attribute editing. Video-P2P\cite{liu2024video} enhances editing through null-text optimization\cite{mokady2023null} but faces leakage with multiple targets. Ground-A-Video\cite{jeong2023ground} introduces box-level attention control, though overlapping boxes lead to texture mixing and detail loss.

\section{Methodology}
\label{sec:method}

Given a series of source video frames $\mathit{f}^{1:N}$ and a text prompt $P$ containing semantic attributes $\{\tau_1, \tau_2, ..., \tau_m\}$, our goal is to accurately edit various attributes of the source video while avoiding unwanted modifications, according to a list of intended edits $\Delta\tau = \{\tau_1\rightarrow\tau_1', \tau_2\rightarrow\tau_2', ..., \tau_m\rightarrow\tau_m'\}$. The target prompt $P'$ is derived from $P$ by applying $\Delta\tau$. Based on the standard noise predictor $\epsilon_\theta$ from Stable Diffusion, we extend its architecture temporally to obtain $\hat{\epsilon_\theta}$ for multi-frame processing. MAKIMA aims to generate an edited video that semantically aligns with $P'$ while preserving the structural composition of the source video.

Our framework consists of three main stages. First, for each attribute to be edited, we obtain its corresponding binary masks $M_m^{1:N}$ through semantic segmentation and tracking using SAM2(\cite{ravi2024sam}). We then perform DDIM inversion on the source video to obtain the initial noise $z_T$ and cache the self-attention maps ($Q_t, K_t$) along with convolutional features $f_t$. The target prompt $P'$ is constructed by applying the attribute edits $\Delta\tau$ to the source prompt $P$.

During the denoising process, we modulate the inflated self-attention using the prepared masks. Specifically, we enhance attention scores between tokens of the same attribute across frames while suppressing different-attribute attention to prevent attention leakage. For cross-attention, we boost the attention scores within each attribute's mask region while attenuating them outside to ensure precise text-to-appearance control. For efficient video generation, our approach edits keyframes and propagates their features to intermediate frames, achieving an optimal trade-off between quality and efficiency.

\begin{figure*}[h]
\centering
\includegraphics[width=\linewidth]{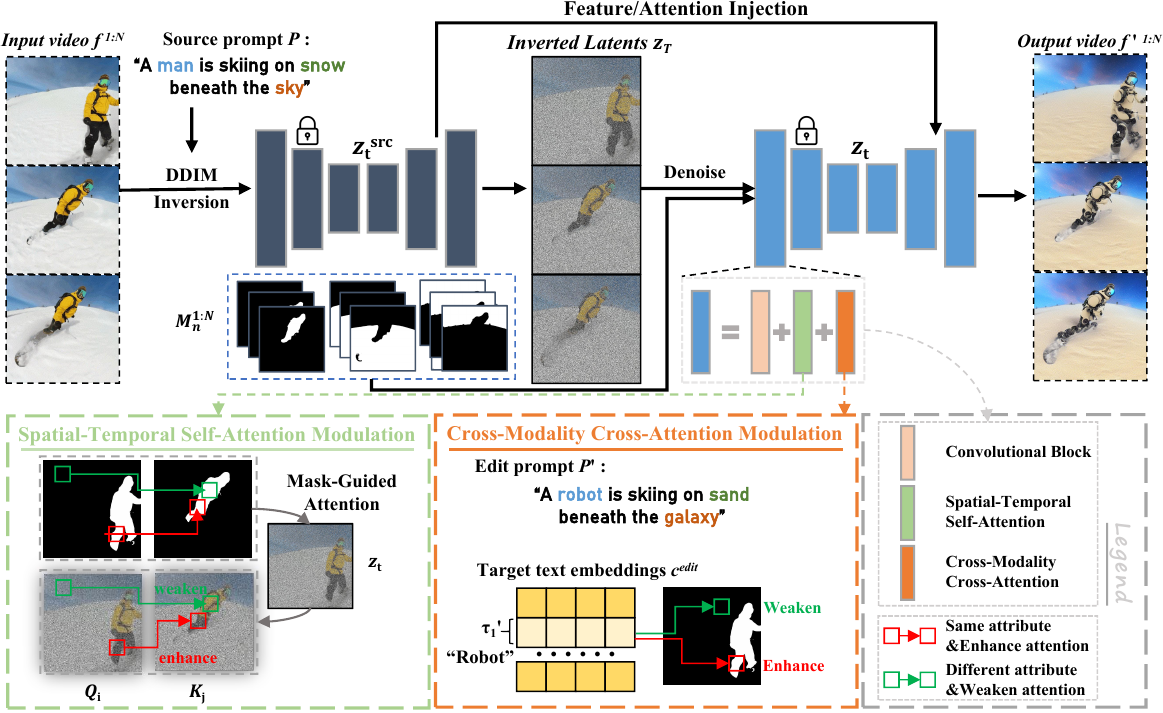}
\caption{MAKIMA pipeline. After performing DDIM inversion to obtain latent features and attention maps, we inflate the UNet for denoising with Mutual Spatial-Temporal Self-Attention. During denoising, we utilize pre-computed attribute masks to guide attention modulation: enhancing intra-attribute correlations while suppressing inter-attribute interference in self-attention, and controlling text-guided appearance transformation in cross-attention.}
\label{fig:method}
\end{figure*}

\subsection{Preliminary for Diffusion Models}

Diffusion models~\cite{ho2020denoising, song2020denoising} are probabilistic generative models that create images through a gradual denoising process. The framework consists of two key phases: a \textit{forward} process that progressively adds Gaussian noise to data, and a \textit{reverse} process that learns to denoise and reconstruct the original data. During generation, the model starts from random noise and iteratively refines it into a high-quality image through learned noise prediction.

Our method builds upon the text-conditioned Stable Diffusion (SD) model~\cite{rombach2022high}, which performs denoising in a compact latent space for efficiency. The pipeline first maps input images into this latent space using a VAE encoder\cite{kingma2013auto}, applies the diffusion process and finally decodes the refined latent back to the image space. In the noise-predicting network $\epsilon_{\theta}$, residual blocks generate intermediate features $f_t^l$, which are further processed by attention mechanisms.

Self-attention captures long-range spatial interactions within the features:
\begin{equation}
    Attention(Q, K, V) = softmax(\frac{QK^\top}{\sqrt{d_k}})V
\end{equation}
where $Q$, $K$, and $V$ represent queries, keys, and values derived from the same feature map, with $d_k$ denoting the key/query dimension.

Cross-attention then integrates the textual prompt $P$ by using it to generate keys and values, merging text and image semantics. These attention layers in the SD model significantly influence image composition and synthesis, guiding editing by manipulating attention during denoising.

\subsection{Mutual Spatial-Temporal Self-Attention}
To achieve temporal consistency, we inflate the spatial self-attention of pretrained text-to-image Stable Diffusion along the temporal dimension without modifying its weights\cite{khachatryan2023text2video, qi2023fatezero}. Specifically, for frame $i$, the query features $Q$ are computed from its latent representation $z_t^i$, while key $K$ and value $V$ features are derived from the concatenated latents $[z_t^1,...,z_t^N]$ across all frames:

\begin{equation}
   Q = W^Q z_t^i, \quad K = W^K z_t^{1:N}, \quad V = W^V z_t^{1:N}
\end{equation}
where $W^Q$, $W^K$, $W^V$ are learnable projection matrices and $z_t^{1:N} = [z_t^1,...,z_t^N]$ represents concatenated frame latents.

To preserve the spatial structure of the source video during editing, we leverage both self-attention maps $Q^{src}_t, K^{src}_t$ and convolutional features $f^{src}_t$ from the DDIM inversion process. The self-attention maps guide the attention distribution of edited content, while the features from conv blocks provide fine-grained structural details. By fusing these spatial cues during denoising, our method effectively maintains the source video's layout while enabling accurate attribute editing. This can be expressed by:
\begin{equation}
    z_{t-1} = \hat{\epsilon_\theta}(z_t, P', t;\{Q^{src}_t, K^{src}_t, f^{src}_t\}) 
\end{equation}

\subsection{Mask-guided Attention Modulation}

For each attribute $\tau_m$ in a frame $i$, there is a corresponding binary mask $M^i_m$ indicating its spatial layout. To achieve accurate text-to-attribute control while preventing attention leakage, we propose a mask-guided attention modulation strategy:
\begin{equation}
A' = softmax(\frac{QK^\top + \Delta_{modu}}{\sqrt{d}})
\end{equation}
where $\Delta_{modu}$ represents our modulation term, defined differently for self-attention and cross-attention layers.

\subsubsection{Self-attention Layer Modulation}
In the self-attention layer, we enhance correlations between tokens from the same attribute $m$ while suppressing inter-attribute interference. For frames $i,j \in [1:N]$, we compute two complementary correspondence matrices:
\begin{equation}
E^m_{p,q} = M^i_m M^j_m
\end{equation}
\begin{equation}
\bar{E}^m_{p,q} = (1-M^i_m)M^j_m
\end{equation}
where $M^i_m$ and $M^j_m$ represent binary masks for attribute $m$ in frames $i$ and $j$ respectively, with $E^m_{p,q}$ identifying token pairs from the same attribute while $\bar{E}^m_{p,q}$ capturing cross-attribute relationships.

Based on this correspondence, we obtain the maximum attention score within the masked region:
\begin{equation}
\alpha_m = \max(Q^i[K^{1:N}]^\top \odot E^m_{p,q})
\end{equation}
where $\odot$ denotes element-wise multiplication.

We then modulate self-attention through:
\begin{equation}
\Delta_{self} = \alpha_m \cdot E^m_{p,q} - \alpha_m \cdot \bar{E}^m_{p,q}
\end{equation}

This enhances attention scores between tokens of the same attribute through the first term while suppressing cross-attribute attention via the second. The modulation strength is adaptively determined by $\alpha_m$, which captures the strongest attention score within the attribute region.

\subsubsection{Cross-attention Layer Modulation}
For the cross-attention layer, we modulate attention between each attribute's spatial region and its corresponding text tokens. For each attribute $\tau_m$, we compute two complementary attention masks:
\begin{equation}
E^m_{p,t} = M^i_m I^{\tau_m}
\end{equation}
\begin{equation}
\bar{E}^m_{p,t} = (1-M^i_m)I^{\tau_m}
\end{equation}
where $M^i_m$ represents the spatial mask for attribute $m$ in frame $i$, and $I^{\tau_m} \in \{0,1\}^L$ is the indicator vector for tokens of attribute $\tau_m$, with $E^m_{p,t}$ identifying spatial-token pairs belonging to the same attribute while $\bar{E}^m_{p,t}$ capturing spatial-token relationships outside the attribute region.

Based on these masks, we obtain the maximum attention score within the masked region:
\begin{equation}
\alpha_m = \max(Q^i[K^{\tau_m}]^\top \odot E^m_{p,t})
\end{equation}

We then modulate cross-attention through:
\begin{equation}
\Delta_{cross} = \alpha_m \cdot E^m_{p,t} - \alpha_m \cdot \bar{E}^m_{p,t}
\end{equation}

This enhances attention between an attribute's text tokens and its masked spatial region through the first term while suppressing attention to other regions through the second term. The modulation strength is adaptively determined by $\alpha_m$, which captures the strongest attention score within the attribute region.

\subsubsection{Regularization}
Since our method alters the original diffusion process, we apply both temporal and spatial regularization to maintain generation quality\cite{kim2023dense}:
\begin{equation} 
\Delta_{modu} = \gamma \cdot \lambda_t \cdot (1-\omega_m) \cdot \Delta_{attn}
\end{equation}
where $\Delta_{attn}$ represents either $\Delta_{self}$ or $\Delta_{cross}$ for respective attention layers. The modulation scale $\gamma$ is set to 0.1 for self-attention and 1.0 for cross-attention. The temporal regularization $\lambda_t = \frac{t}{1000}$ reduces modification strength as $t$ approaches zero to prevent quality degradation in later denoising steps. The spatial regularization $\omega_m = \frac{\sum_p M^i_m[p]}{|V|}$ adaptively adjusts modulation intensity based on the ratio of mask area to total spatial dimension $|V|$, preventing over-modification in regions with significant size differences.

\subsection{Consistent Feature Propagation}
Cross-frame editing with attention modulation significantly increases computational overhead. To balance quality with computational efficiency, we perform detailed editing with mask-guided self-attention modulation on perceptually significant keyframes, then propagate their modulated self-attention features to neighboring frames for temporal consistency. For a video with $N$ frames, we select a set of $k$ keyframes $s_1$ that satisfies:
\begin{equation}
s_1 = \{f_{i_1},...,f_{i_k}\} \text{ s.t. } \min_{i,j \in s_1, i \neq j} D(f_i, f_j) \text{ is maximized}
\end{equation}
where $D(f_i, f_j)$ measures the cosine distance between frame features. This formulation ensures that selected keyframes capture significant perceptual variations in the sequence. We implement the selection process using binary search to determine optimal distance thresholds. During the denoising process, we adaptively combine $s_1$ with supplementary frames to maintain consistent feature propagation.
For non-keyframes, we introduce a content-aware weighting scheme that prioritizes feature similarity while considering temporal relationships:
\begin{equation}
A'_{self,i} = w_1 A'_{self,k1} + (1-w_1)A'_{self,k2}
\end{equation}
where $w_1$ is computed through a consistency-oriented weighing mechanism:
\begin{equation}
w_1 = \sigma(\frac{w_{temp} \cdot sim_1}{w_{temp} \cdot sim_1 + (1-w_{temp}) \cdot sim_2})
\end{equation}
Here, $w_{temp} = \frac{d_2}{d_1 + d_2}$ incorporates temporal context based on distances $d_1, d_2$ to adjacent keyframes, while $sim_1, sim_2$ measure the cosine similarities between the non-keyframe features and their corresponding keyframe features. This sophisticated weighting scheme ensures smooth and consistent feature propagation across the video sequence while maintaining temporal coherence.

\section{Experiments}
\label{sec:experiments}

\begin{figure*}[t]
\centering
\includegraphics[width=0.98\linewidth]{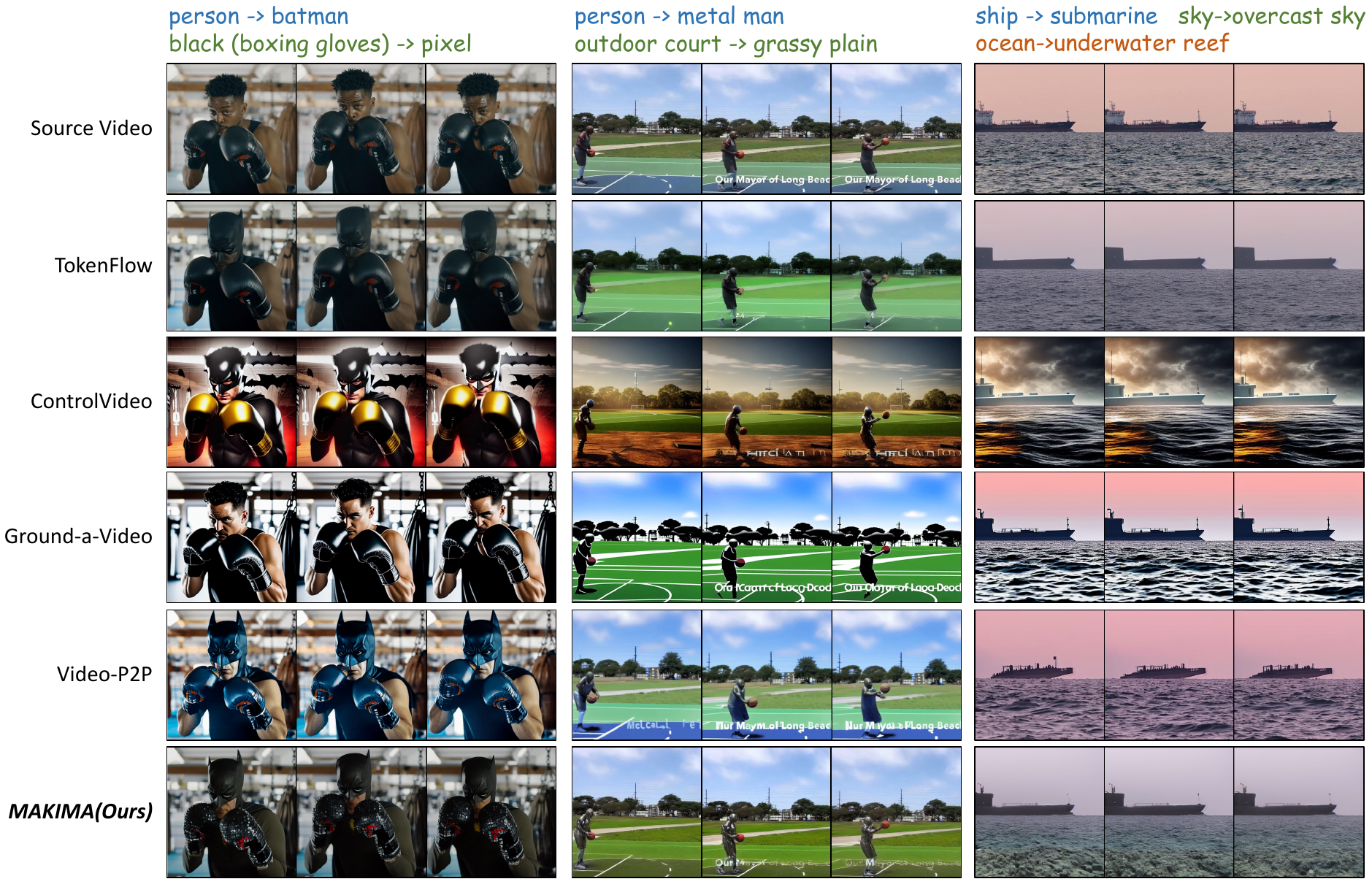}
\caption{Qualitative comparison with baselines: Our method achieves precise attribute-specific editing while maintaining structural consistency with the original video frames.}
\label{fig:comp}
\end{figure*}

\subsection{Implementation Details}

We evaluate our framework on a diverse set of 33 videos from DAVIS dataset\cite{perazzi2016benchmark} and the Internet, with manually annotated text descriptions. For each video sequence, we generate three distinct text prompts targeting multiple attribute modifications, yielding 99 video-text pairs for evaluation. We extract 12 frames from each video and process them to a resolution of 512$\times$512. To obtain attribute masks efficiently, we implement an automated pipeline: GroundingDINO\cite{liu2023grounding} detects regions of interest based on attribute-specific prompts, followed by SAM2\cite{ravi2024sam} for precise mask generation within the detected regions.

We adopt the pre-trained Stable Diffusion v1.5\cite{rombach2022high} as our base model. The video editing pipeline initializes with a 50-step DDIM deterministic inversion\cite{song2020denoising} to obtain latent representations for each frame. During inference, we implement 50-step DDIM sampling with a classifier-free guidance scale of 7.5\cite{ho2022classifier}. The proposed Mask-guided Attention Modulation is applied globally across all network layers throughout the sampling process. For mutual self-attention, we conduct self-attention injection in decoder layers 4-11 and feature injection in decoder layer 4 during the first 25 and 40 denoising steps, respectively. To optimize efficiency, we sample three keyframes per denoising step and propagate their features across the sequence. Both Mutual Spatial-Temporal Self-Attention and Self-attention Layer Modulation are applied to keyframes only. All experiments are conducted on a single NVIDIA L20 GPU.

\subsection{Comparison}

\textbf{Metrics.} We evaluate our method using four complementary metrics to assess both editing quality and temporal consistency as \cite{geyer2023tokenflow, yang2024fresco}:
\textit{CLIP-T} calculates the average cosine similarity between the edited frames and the target text prompt in CLIP feature space, quantifying how well our edited video semantically aligns with the intended modifications.
\textit{CLIP-F} measures the temporal coherence by computing the average cosine similarity between adjacent frames in CLIP feature space. This metric helps evaluate whether our editing maintains a consistent visual appearance across frames.
\textit{Frame Acc} assesses editing effectiveness by calculating the percentage of frames where the CLIP similarity to the target prompt surpasses that to the source prompt, indicating successful attribute transformation.
\textit{Runtime} evaluates computational efficiency by measuring the total processing time required for video editing, including model fine-tuning.

\noindent \textbf{Baselines.} 
(1) TokenFlow\cite{geyer2023tokenflow} achieves tuning-free video editing by jointly editing sampled keyframes at each denoising step and propagating features through linear weighting; (2) ControlVideo\cite{zhang2023controlvideo} adopts a tuning-free approach that feeds source video information into ControlNet and leverages cross-frame attention to generate edited videos; (3)Ground-A-Video\cite{jeong2023ground} optimizes null-text embeddings and employs ControlNet guidance with attribute-specific bounding boxes for multi-attribute video editing; (4)Video-P2P\cite{liu2024video} optimizes shared unconditional embeddings for the editing branch and achieves attribute-specific editing through cross-attention control.

\noindent \textbf{QUALITATIVE EVALUATION.} Figure \ref{fig:comp} demonstrates the effectiveness of various video editing methods. In the left person-centric example, existing methods fail to transform ``black boxing gloves'' to ``pixel boxing gloves'' due to insufficient alignment between spatial regions and text embeddings in their attention mechanisms, while MAKIMA achieves accurate editing through enhanced cross-attention in specific regions. The middle outdoor sports scene reveals multiple limitations: TokenFlow suffers from information loss in the subject's legs due to watermark interference disrupting feature propagation and showing degraded background preservation. ControlVideo's depth map-based generation completely fails to preserve background details and incompletely transforms the ``grassy plain''. Ground-A-Video generates semantically correct content but loses substantial visual details due to overlapping attribute bounding boxes interfering with cross-attention modulation. Video-P2P's limited cross-attention control overlooks the ``grassy plain'' modification while focusing solely on the ``metal man'' transformation. In the right ship-to-submarine example, all baseline methods fail to properly transform the ``ocean'' into ``underwater reef'' due to their limited capability in handling concurrent attribute modifications. Additionally, TokenFlow completely loses ``submarine'' details during feature propagation, while Video-P2P fails to preserve the ship's structural integrity during the ship-to-submarine transformation. These results demonstrate the importance of precise attention control in multi-attribute video editing, which MAKIMA achieves through its mask-guided attention modulation approach.

\noindent\textbf{QUANTITATIVE EVALUATION.}
\textbf{Automatic Metrics.} We evaluate our method against the baselines using automatic metrics, with results presented in Table \ref{tab:comp}. Our method achieves superior performance on CLIP-T and Frame-Acc. Although TokenFlow demonstrates impressive frame continuity, its consistency largely stems from conservative editing rather than effective feature propagation. While it successfully propagates low-variation features between frames, it struggles to implement more substantial prompt modifications - as evidenced by its significantly lower Frame-Acc scores compared to our method. Meanwhile, ControlVideo's high CLIP-F scores obscure a critical weakness: its ControlNet-based generation tends to oversimplify background details, resulting in visually static background objects. This loss of background dynamics artificially inflates temporal consistency metrics without preserving the temporal richness of the original video. Ground-A-Video struggles with its cross-attention modulation when handling overlapping attribute regions, as the bounding boxes of edited elements frequently intersect. Similarly, while Video-P2P performs well in single-target editing scenarios, its attention control mechanism has difficulty managing multiple elements within the same video frame. Our method achieves substantial improvements over prior tuning-free approaches while maintaining reasonable computational overhead, significantly lower than methods requiring fine-tuning, demonstrating an effective balance between performance and efficiency.

\begin{table}[ht]
\centering
\setlength{\tabcolsep}{4pt}
\caption{Quantitative comparison with state-of-the-art methods. $\uparrow$ indicates higher is better, $\downarrow$ indicates lower is better.}
\resizebox{0.45\textwidth}{!}{%
\renewcommand{\arraystretch}{1.1}
\begin{tabular}{@{}l|cccc@{}}
\toprule[2pt]
\textbf{Method} & \textbf{CLIP-T} $\uparrow$ & \textbf{CLIP-F} $\uparrow$ & \textbf{Frame-Acc} $\uparrow$ & \textbf{Runtime} $\downarrow$ \\
\midrule[1.5pt]
ControlVideo & 0.3304 & \textbf{0.9965} & 92.57\% & \textbf{45s} \\
TokenFlow & 0.3312 & 0.9852 & 83.91\% & 56s \\
Ground-A-Video & 0.3202 & 0.9735 & 75.72\% & 143s \\
Video-P2P & 0.3188 & 0.9664 & 75.49\% & 260s \\
\rowcolor{blue!5} \textbf{\textit{MAKIMA}}(Ours) & \textbf{0.3387} & 0.9799 & \textbf{98.65\%} & 65s \\
\bottomrule[2pt]
\end{tabular}
}
\label{tab:comp}
\end{table}

\begin{figure}[ht]
\centering
\includegraphics[width=\linewidth]{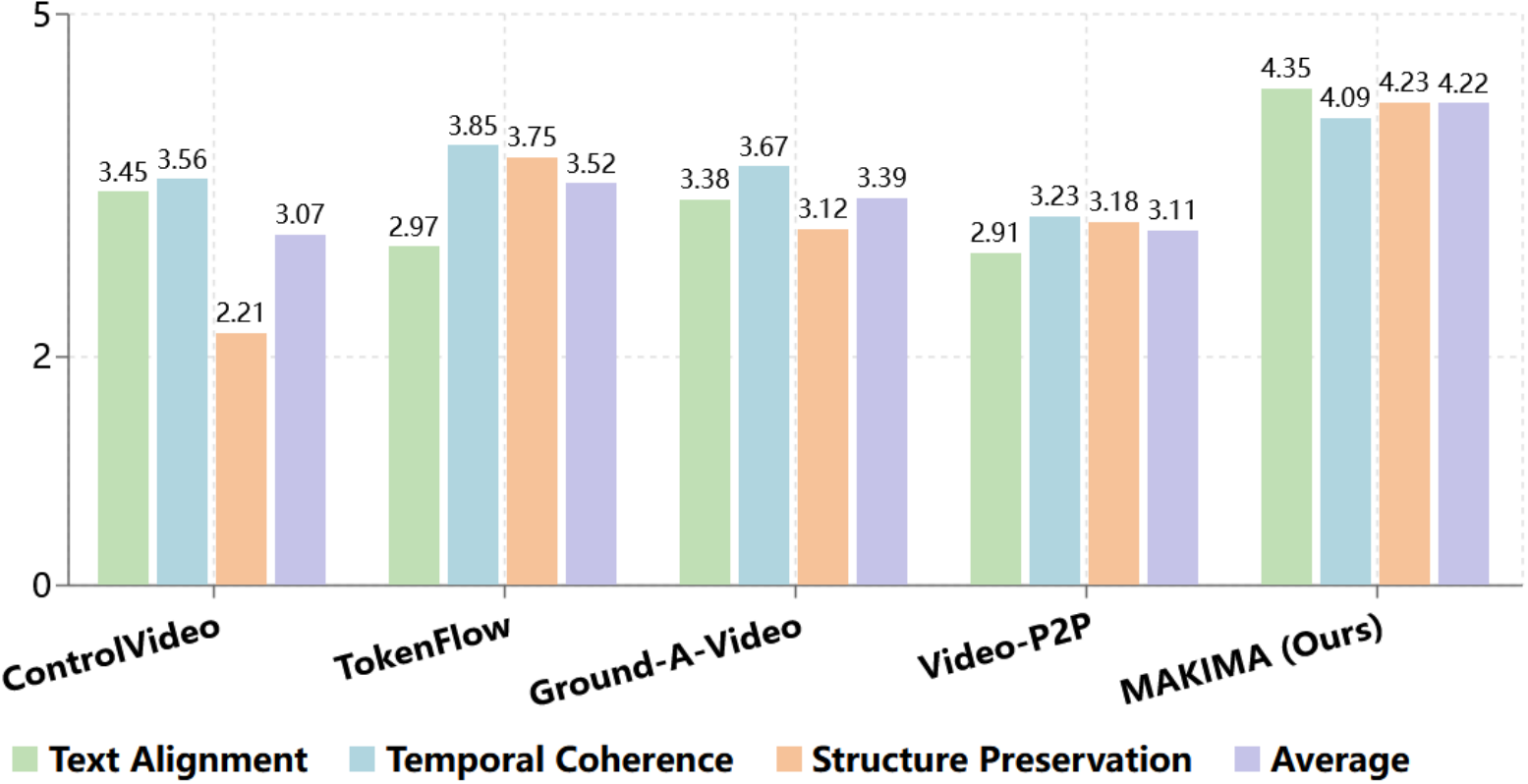}
\vspace{-2mm}
\caption{User study results comparing different methods.}
\label{fig:user}
\end{figure}

\noindent\textbf{User Study.} 
To further validate the effectiveness of our approach, we conduct a comprehensive user study evaluating edited video quality across three key aspects: (1) Text Alignment (T-Align) - the degree of semantic consistency with the edit prompt, (2) Temporal Coherence (T-Coh) - the smoothness and continuity between video frames, and (3) Structure Preservation (S-Pres) - the degree of structural similarity with the source video, particularly in unedited regions.
The study involves 25 participants who rate videos using a 5-point Likert scale. As shown in Figure \ref{fig:user}, our method achieves superior performance across all metrics. Notably, our approach demonstrates significant advantages in Text Alignment compared to ControlNet-guided methods like ControlVideo and Ground-A-Video, underscoring the effectiveness of our mask-guided attention modulation strategy in achieving precise attribute-specific editing.

\subsection{ABLATION STUDY}
\textbf{Attention.} We conduct ablation studies on three key components: Mutual Spatial-Temporal Self-Attention (MSTA), Mask-guided Attention Modulation (MAM), and Consistent Feature Propagation (CFP). 

\begin{figure}[ht]
\centering
\includegraphics[width=1\linewidth]{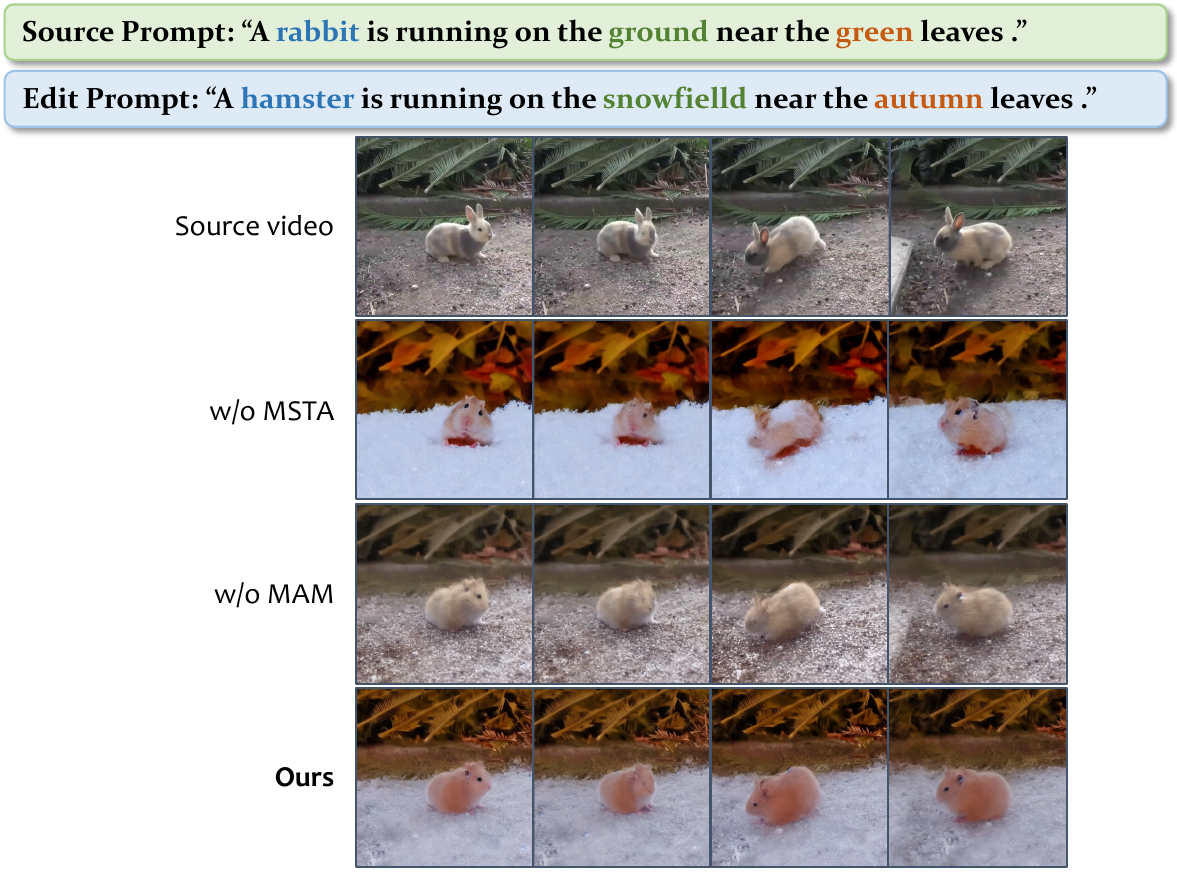}
\caption{\textbf{Ablation study.} Without MSTA, the generated video frames fail to maintain temporal and spatial consistency with the source video. In the absence of MAM, the generated video lacks the capability to achieve desired attribute-specific modifications.}
\label{fig:ablation}
\end{figure}

\begin{table}[t]
\centering
\small
\setlength{\tabcolsep}{4pt}
\caption{Quantitative ablation of key components of MAKIMA.}
\resizebox{0.38\textwidth}{!}{%
\renewcommand{\arraystretch}{1.1}
\begin{tabular}{l|cccc}
\toprule[2pt]
\textbf{Method} & \textbf{CLIP-T} & \textbf{CLIP-F} & \textbf{Frame-Acc} & \textbf{Runtime} \\
\midrule[1.5pt]
w/o MSTA & 0.3396 & 0.9692 & 99.16\% & 41s \\
w/o MAM & 0.3278 & 0.9849 & 83.33\% & 58s \\
w/o CFP & 0.3396 & 0.9767 & 98.74\% & 217s \\
\rowcolor{blue!5} \textbf{\textit{Ours}} & 0.3387 & 0.9799 & 98.65\% & 65s \\
\bottomrule[2pt]
\end{tabular}
}
\label{tab:ablation}
\vspace{-2mm}
\end{table}

\noindent \textbf{Mutual Spatial-Temporal Self-Attention (MSTA).}  
Through ablation studies (Figure \ref{fig:ablation}), removing MSTA significantly impacts visual coherence and continuity. While this leads to higher CLIP-T and Frame-Acc scores by generating content more aligned with prompts, it results in inconsistent ``rabbit'' appearances and poor preservation of autumn leaves' structure, reflected in lower CLIP-F scores. These results demonstrate MSTA's importance in maintaining both temporal dynamics and spatial fidelity.

\noindent \textbf{Mask-guided Attention Modulation(MAM).}  
MAM serves as the cornerstone of our approach for multi-attribute video editing. As shown in Figure \ref{fig:ablation} (comparing the third and last rows), the absence of MAM significantly impairs the model's ability to execute desired attribute modifications. Without MAM, the model fails to correctly edit key attributes. While this configuration tends to produce temporally coherent frames, its inability to properly respond to attribute-specific editing prompts results in decreased CLIP-T and Frame-Acc scores. These observations validate the effectiveness of our MAM component in achieving precise multi-attribute editing while maintaining content integrity in unmodified regions.

\subsection{Regularization on Attention Modulation}
\begin{figure}[ht]
\centering
\includegraphics[width=\linewidth]{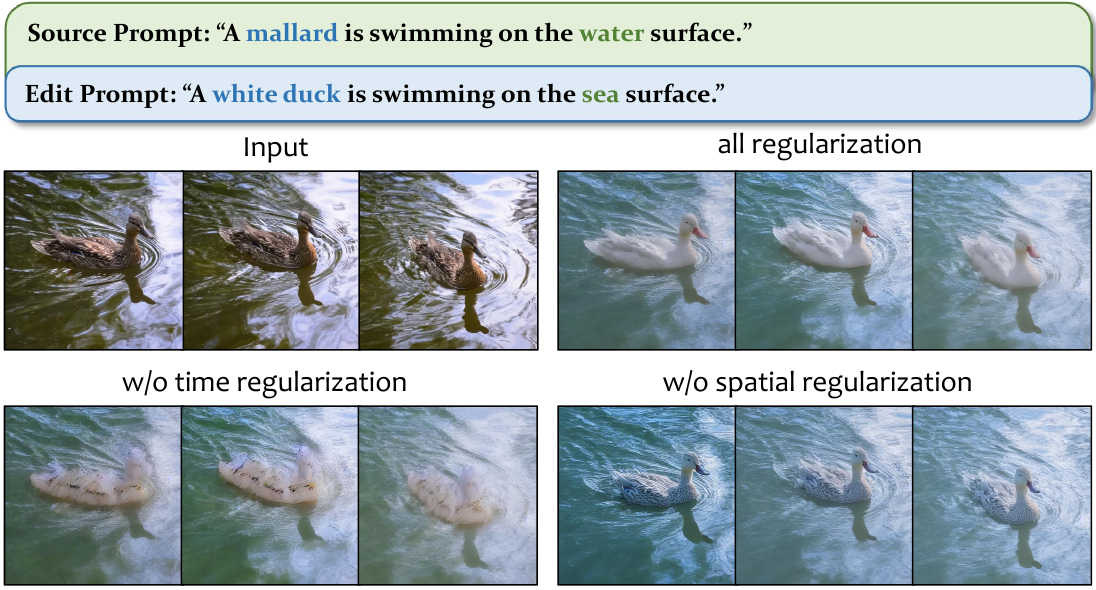}
\caption{Ablation Study on Modulation Regularization. Temporal regularization ensures gradual reduction of attention modulation during denoising to prevent video quality degradation. Spatial regularization balances editing intensity across multi-scale attributes. }
\label{fig:regular}
\end{figure}

As the denoising process progresses, the model generates increasingly distinct features. However, employing a fixed attention modulation strategy may lead to the generation of low-quality images as shown in Fig. \ref{fig:regular}. To address this issue, we introduce temporal regularization to attenuate the modulation intensity at later denoising steps.
Significant disparities in attribute areas can result in imbalanced modulation during attention modulation, as illustrated in Fig. \ref{fig:regular}. In this example, while the ``sea'' attribute is effectively edited, the ``white duck'' does not receive comparable modulation intensity, resulting in inconsistent editing effects. 

\section{Conclusion}
\label{sec:conclusion}
We propose MAKIMA, a tuning-free framework that enables precise and localized editing of arbitrary attributes in videos. 
The Mutual Spatial-Temporal Self-Attention is devised to maintain structural consistency between generated frames and the source video, and leverage Mask-guided Attention Modulation to regulate attention distributions, enhancing intra-attribute feature correlations while suppressing inter-attribute interference. Experiments validate the superiority of MAKIMA.
{
    \small
    \bibliographystyle{ieeenat_fullname}
    \bibliography{main}
}


\end{document}